\documentclass[letterpaper, 10 pt, conference]{ieeeconf}  
\IEEEoverridecommandlockouts                              % This command is only
\usepackage{graphicx} % for pdf, bitmapped graphics files
\graphicspath{ {./images/} }
\usepackage{rotating}
\usepackage{float}
\usepackage{lipsum}

\usepackage{afterpage}

\title{\LARGE \bf
A Persistent and Context-aware Behavior Tree Framework for Multi Sensor Localisation in Autonomous Driving}

\author{Siqi Yi, Stewart Worrall and Eduardo Nebot*% <-this % stops a space
\thanks{*Authors are from Intelligent Transport Systems group, Australian Center for Field Robotics, the University of Sydney, Australia.
E-mails: {\tt\small yisiqiyisiqi@gmail.com, \{s.worrall, e.nebot\}@acfr.usyd.edu.au}}% <-this % stops a space
}

\begin{document}
\maketitle
\thispagestyle{empty}
\pagestyle{empty}

%%%%%%%%%%%%%%%%%%%%%%%%%%%%%%%%%%%%%%%%%%%%%%%%%%%%%%%%%%%%%%%%%%%%%%%%%%%%%%%%
\begin{abstract}

% why switching/turn on and off? why not tune up and down weights of good and bad sensors? because GPS often has a constant bias. GPS noise are highly non parametric and non Gaussian. fusing such noises are challenging. This is especially true when higher accuracy localisation method are available, we are better off turning off non parametric and noisy update sensor sources.

% relay running of sensors
% Localisation systems for autonomous driving have utilised direct or feature based visual and lidar methods, radar, GPS, as well as using a combination of such sensors and methods. 
% In addition to real time health metrics, the behavior tree also base its context aware decisions making process on location based information that are computed and stored in spatial map database prior to real time localisation. %The framework was implemented in ROS and achieved persistent localisation in the University of Sydney dataset, in which an electrical vehicle fitted with multiple sensors collected 53 rosbags in a diverse environment for a total of over 227km  during 1.5 years.
Robust and persistent localisation is essential for ensuring the safe operation of autonomous vehicles. When operating in large and diverse urban driving environments, autonomous vehicles are frequently exposed to situations that violate the assumptions of algorithms, suffer from the failure of one or more sensors, or other events that lead to a loss of localisation. 

This paper proposes the use of a behavior tree framework that can monitor the performance of localisation health metrics and triggers intelligent responses such as sensor switching and loss recovery. The algorithm presented selects the best available sensor data at given time and location, and can perform a series of actions to react to adverse situations. 
The behavior tree encapsulates the system-level logic to give commands that make up the intelligent behaviors, so that the localisation ``actuators" (data association, optimisation, filters, etc) can perform decoupled actions without needing context. Experimental results to validate the algorithms are presented using the University of Sydney Campus dataset which was taken weekly over an 18 month period. A video showing the online localisation process can be found here: https://youtu.be/353uKqXLV5g

\end{abstract}

%%%%%%%%%%%%%%%%%%%%%%%%%%%%%%%%%%%%%%%%%%%%%%%%%%%%%%%%%%%%%%%%%%%%%%%%%%%%%%%%

\begin{figure}
    \centering
    \includegraphics[width=\columnwidth]{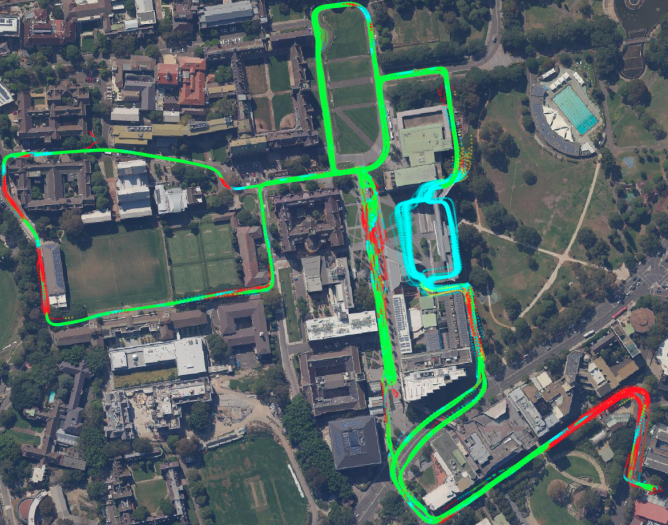}
    \caption{Vehicle trajectories of 53 rosbags in the University of Sydney Campus Dataset output from the GPS+lidar main filter are superimposed together onto a satellite image. No manual alignment of trajectory is performed. The color of trajectory points indicates the state of update used by the filter. Green: lidar feature updating. Blue: dead reckoning. Red: GPS updating. The blue dead reckoning area in the middle of the image is an underground carpark.}
    \label{fig:lidar_traj}
\end{figure}

\section{INTRODUCTION}

Robust and persistent localisation is an essential task for autonomous driving. To guarantee driving safety when operating with high levels of autonomy, localisation systems must be able to achieve a certain accuracy within a large and diverse urban driving environment all of the time and anywhere, without the need for human intervention. Current localisation solutions come up short of such requirements in many of the following aspects. 

First, the performance of available single sensor localisation solutions are not robust against diverse environments and changing situations. Changes of the environment often cause failure of sensor models or changes to sensor noise profiles. For example, feature based localisation methods achieve high accuracy in feature-dense areas, but will be lost in areas where features are ill-formed or sparse. GPS measurements can be affected by multi-path errors and report large biases. 
% For a single sensor modality, . 

\begin{figure}
    \centering
    \includegraphics[width=\columnwidth]{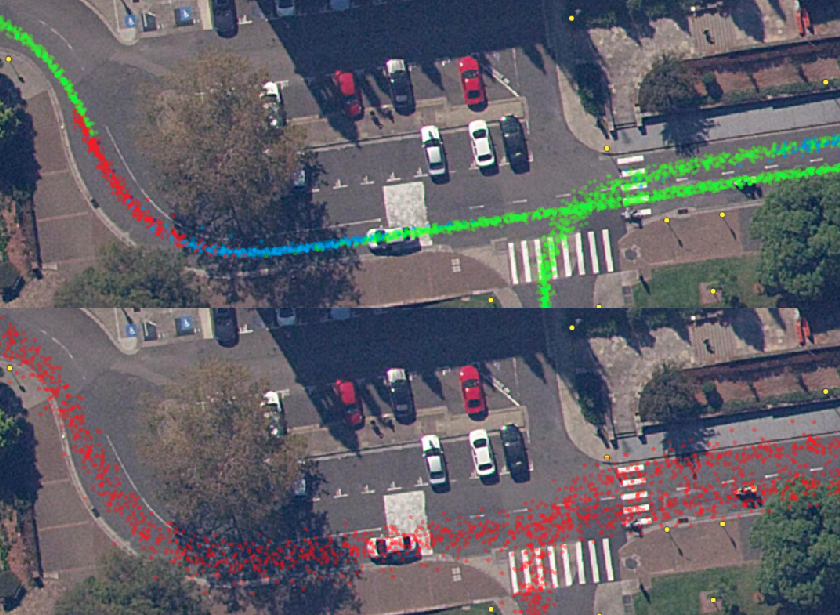}
    \caption{Zoomed in trajectories of 53 rosbags at a location where transitions are clear and consistent. Top: GPS+lidar main filter. Bottom: backup GPS filter. Green: lidar feature updating. Blue: dead reckoning. Red: GPS updating. Yellow: lidar map features. }
    \label{fig:transition_zoomin}
\end{figure}

Exploiting multi sensor redundancy has become a norm in autonomous driving, since each sensor modality provides resilience to different environments and conditions. However, due to calibration errors, imperfections in the map making process, and the different accuracy of different sensors, there often exists biases between sensor modalities. The biases and errors can not be easily characterised using current frameworks. 
% This problem can be addressed with multiple sensors. 
One of the fundamental challenges of sensor integration is how to choose the best sensor combinations, and how to determine the priorities or weightings for each sensor modality.

Due to the complexity of the driving environment and localisation algorithm performances, autonomous vehicles inevitably encounter occasions when none of the sensors works well, where outliers and errors remain undetected, or when unpredictable edge cases occur in the environment. In real urban driving scenarios, the addition of these factors could lead to the loss of localisation.
% that should be expected to happen more regularly than rarely.  

% Localization estimations are often needed as an input for system level decision making algorithms such as path planning and obstacle avoidance. However, because of the many challenges introduced by the diverse and changing urban driving environment, the ability of context aware intelligent localisation behaviors that can detect and respond to dynamic situation and environmental changes properly. Decision making algorithms that can detect situation changes and initiate falling back and failure recovery actions is an essential component of any localisation algorithm to guarantee a given performances. 

To achieve the robustness requirement for autonomous driving, localisation systems should employ context-aware intelligent behaviors to address these challenges. First, localisation systems need to be aware of environmental changes and performance changes of every on-board sensor while driving in a diverse urban environment. Then the localisation systems should possess the ability to make appropriate decisions as to whether actions need to be taken to respond to the change of situation and which actions should be applied. Lastly, the localisation systems should be able to carry out the actions and coordinate the state transitions caused by the actions.
% Fig.\ref{fig:changes} summarized some of these behaviors. BTs are well suited for the systematic handling of these behaviors.
% each of these behaviors may contain a sequence of decisions being made by the behavior tree and a corresponding sequence of individual actions performed by the data processing modules.

In this paper, we use Behavior Trees (BT) to achieve above mentioned functionalities systematically. Fig. \ref{fig:lidar_traj} and Fig. \ref{fig:transition_zoomin} present an example of how the system operates. In the top image of Fig. \ref{fig:transition_zoomin}, a vehicle was driving from the right side to the left side of the image. Lidar features are densely clustered at both edges of the image, but are very sparse in the middle section. GPS is available throughout the trajectory, but is less accurate than the lidar feature based localisation. The BT localisation system triggered transitions based on the environment and the available sensor algorithms. The first part of the trajectory used lidar feature updates, and then dead reckoned when features became sparse. GPS updates are used for a short period, and system switched back to using lidar features when the environment again contains rich lidar features. Fig. \ref{fig:lidar_traj} present a much more comprehensive example covering many different scenarios corresponding to the campus of the University of Sydney taken over an 18 month period. The figure shows the consistent transitions of localisation sensors across different scenarios.

BT is a system level tool to coordinate and execute tasks and actions. It allows logical transitions to be scaled up in a systematic and modular way. We propose a BT-based localisation framework that cleanly separates decision making and data processing into 2 different programs. The first is the data processing program that we call ``localiser", introduced in section \ref{localiser}. The second program is the decision making program using BT, detailed in section \ref{bTree}. An implementation of such a BT localisation framework based on GPS, lidar and dead reckoning is described in section \ref{experiments} and evaluated in section \ref{result}. A comprehensive literature review is presented in section \ref{relatedwork}.

The BT framework comes with many benefits.
Modern localisation systems are often comprised of many modules and threads arranged in complex pipelines. The addition of intelligent behaviors to these tightly weaved complex systems makes state transition hard to track and manage. The BT framework addresses this problem by separating the decision making and data processing tasks. This separated structure also requires both the localiser and the BT algorithms to be flexible, modular and reusable in different vehicle sensor setups and map environments. 

An additional benefit is that the state and actions of the localisation system are transparent and explicit. The BT framework provide a user friendly graphical interface to notify human operator/driver the status, stage and health of localisation. It also presents a clear interface to inform downstream programs the status and health of the localisation outcomes.

\section{RELATED WORK} \label{relatedwork}

\subsection{Behavior trees(BTs)} \label{bt_intro}
Behavior trees was originally used in the gaming industry to model the behaviors of non-player characters. Since then, it has been adopted to robotic planning and control. It provides modular, hierarchical and flexible control structure to react to dynamically changing situations. A comprehensive introduction and survey to BTs can be found in \cite{colledanchise2018behavior}
and \cite{iovino2020survey}.

BTs have been widely used in robotic ground vehicle navigation.
ROS2 navigation2 package \cite{macenski2020marathon} used BTs to successfully navigate through a crowded building using 2 types of ground robots.
The Move Base FLex ROS package \cite{putz2018move} are deployed in Magazino \cite{magazino_gmbh_2020} warehouse robots and used BTs to navigate 6 commercial settings.
In \cite{olsson2016behavior} a simulated autonomous driving system for behaviors such as overtaking and obstacle avoidance is presented.
BTs are also utilised to control UAVs and swarms of UAVs \cite{goudarzi2020uav}.
\cite{Lan2019ICCV} used BTs for intermediate level control to execute task planning and actuating of micro aerial vehicles. BTs are used for drone inspection in congested traffic environments in \cite{Goudarzi2019} and completed trial runs at the Clifton Suspension Bridge. Other applications include multi agent underwater robots \cite{ozkahraman2020combining} and mobile manipulators for picking\cite{zhou2019autonomous} and cleaning households\cite{french2019learning}.

In contrast to all the above mentioned works that use BT for robotic planning and navigation, this paper applies BT to the robot localisation task.

\subsection{Intelligent behaviors in robot localisation}
To the best of the authors' knowledge, BT have not been used to supervise and control robotic perception and localisation tasks. However, there have been attempts to detect and react to dynamic situations using other mechanisms.

\cite{ferreira2019loose} proposed a decision-making algorithm about the usability of Ultra-wideband radars measurements based on the identification of non-line-of-sight conditions.

\cite{liu2020method} suggests a methodology to generate a tolerance boundary of vehicle position using static ground truth environmental objects. If a monitoring software detected that the real time localization estimation went beyond this boundary, a vision based failure recovery action would be triggered.

Robust graph SLAM backends such as dynamic covariance scaling\cite{agarwal2013robust} dynamically scales down the influence of outlier observations. However this process happens in the background black box of graph optimisation. The scale factor changes are implicit and hard to retrieve.

Variants of visual inertial SLAM such as Orbslam3\cite{campos2020orb} implement various robustness strategies. A multi-map architecture is used in \cite{campos2020orb} so that a loss of visual tracking only results in the current submap being archived, and tracking continues in a new submap. Archived submaps can be recovered when the mapped areas are revisited. When not enough feature matches are detected, \cite{campos2020orb} also performs a wider baseline search in order to maintain tracking. Although greatly improved robustness, these techniques often come with the drawback of scattering logical conditions and control parameters across multiple code modules. The transitions happen under the hood without being systematically reported and managed.    

A line of research uses Active Localisation to improve localisation performance by moving robots to advantageous feature rich locations or poses within an environment to avoid featureless areas. This allows the robot to perceive features or other environmental information more continuously to improve localisation. \cite{zhang2019beyond} proposes active visual localisation to plan motion and optimise for the best localisation accuracy. \cite{rodrigues2018low} generates a control action that drives the vehicle towards the goal, while still favoring feature rich areas within a local scope, thus improving the localization performance. In \cite{zhang2020connectivity}, multi follower robots cooperatively plan their paths to minimize the localisation uncertainty of the leader robot.

The active localisation strategies are not usually effective in typical urban scenarios. In crowded urban environments, the drivable area is usually restricted compared to aerial vehicle applications. For example, vehicle maneuvers within a constrained road usually generally do not change the set of visible features or satellites visibility. Furthermore, in wider multi-lane roads, the action of changing lane is restricted by the driving situation, such as the motion and intention of surrounding vehicles, pedestrians, and traffic rules. In the proposed method of this paper, instead of moving in the physical space, the actions are executed within the localisation software. Some examples of such actions could be enabling GPS updates, determining if lidar-based localisation is lost, or performing a series of actions to recover a lost visual tracking thread.

\section{The localiser: dynamically re-configurable localisation data processing pipelines}\label{localiser}

\begin{figure}
    \centering
    \includegraphics[width=\columnwidth]{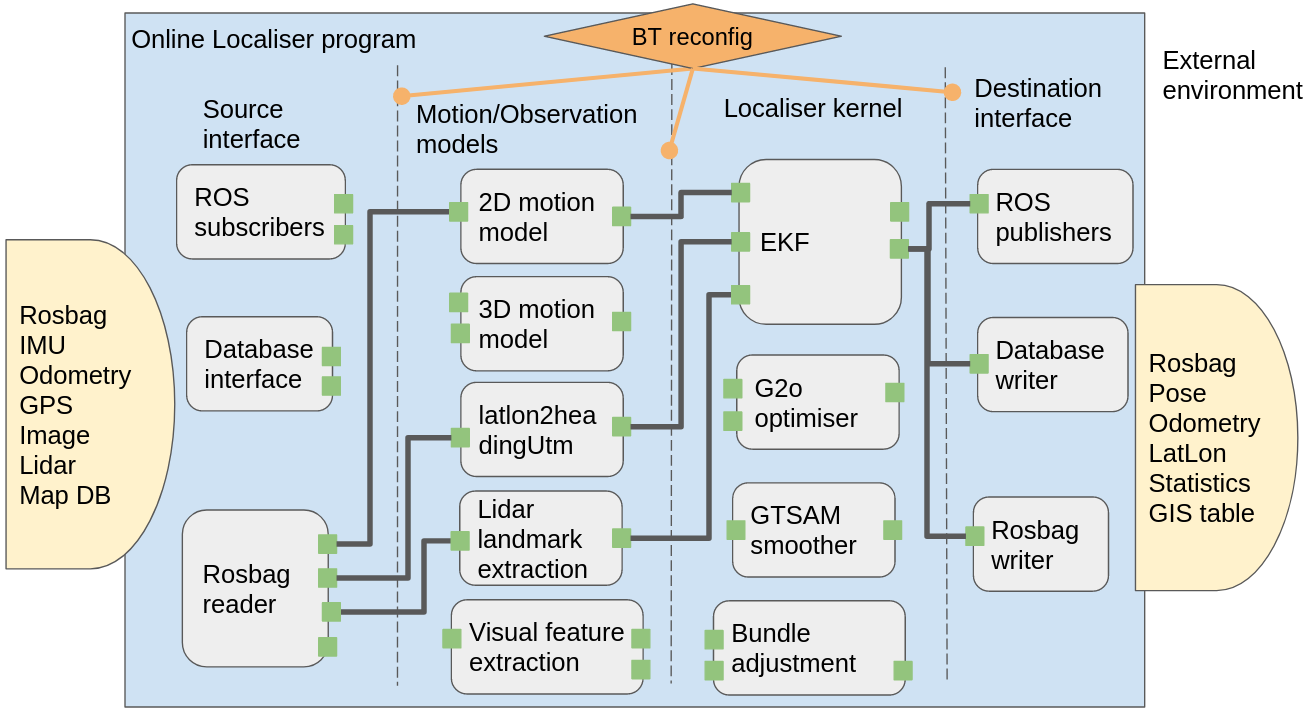}
    \caption{The modular design of the localisation data processing program(localiser)}
    \label{fig:localiser}
\end{figure}

The program for data processing (referred to in this paper as ``localiser" for convenience), consists of the required modules for traditional localisation pipelines such as feature extraction, data association, outlier rejection, filtering and optimisation algorithms for various sensor/feature types, state spaces and processing methods. Some examples of these modules are represented as grey blocks in Fig. \ref{fig:localiser}.

The modules are loosely categorized and organized into 4 layers, namely source interface, sensor/motion models, localisation kernels, and destination interface. The 3 grey dashed lines divided the 4 layers and represent the 3 interfaces where external programs such as BT can reconfigure the inter-module connections on the fly. The input and output of the localiser program is shown in yellow blocks. The input of source interface layer are sensor measurements received from either upstream programs or different formats of storage such as databases and rosbags. The destination interface layer formats the localisation results and output them to the external software environment. 

\begin{figure}
    \centering
    \includegraphics[width=\columnwidth]{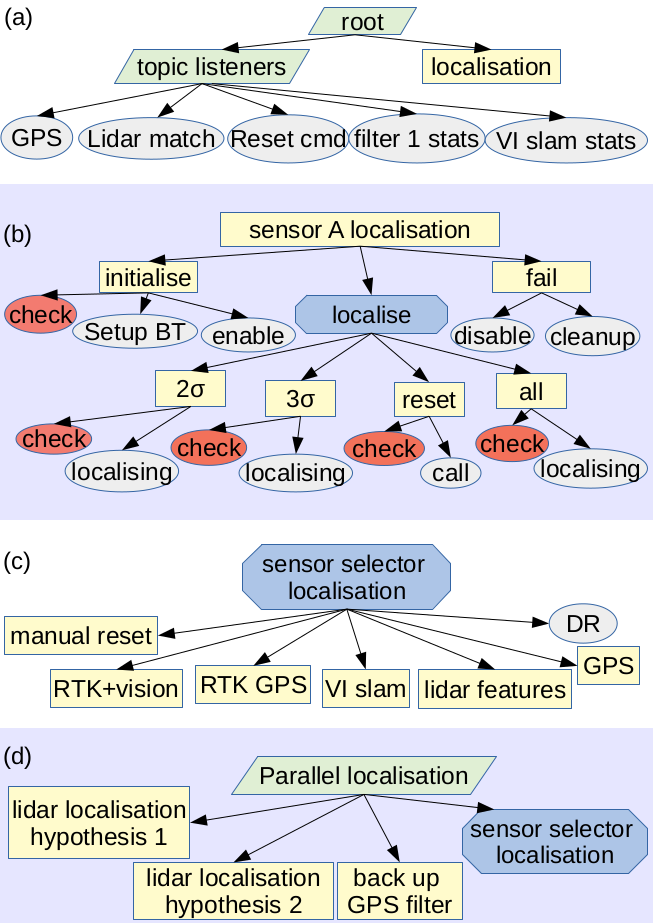}
    \caption{Proposed structures of behavior trees(BTs) to (a) gather contextual information and localiser performance (b) facilitate state transition and failure recovery of single sensor modality localisation (c) fall back to another localisation sensor, feature type or combinations of modalities (d) in parallel run a number of light-weight localisation hypothesis and fall-back options. Light yellow rectangles: Sequence nodes. Blue octagons: selector nodes. Green parallelograms: parallel nodes. Red ellipses: condition nodes. Grey ellipses: actions.}
    \label{fig:bt_models}
\end{figure}

In a behavior tree supervised framework, in order for the decision making program to have full control of the localiser, the building block modules are required to be decoupled from each other and reusable. They can be assembled into multiple data processing pipelines to suit different on-board sensor setups and diverse physical environments. The reusability and modularity also enable external programs to dynamically reconnect and chain the modules together on the fly. During online localisation, the same streams of sensor data flow through different newly formed pipelines that consists of different modules and perform different tasks.

The output of each module can be connected to multiple inputs of following modules to form data processing pipelines. This can be achieved because the module output is type matched to the input of subsequent modules. We implemented this connect/reconnect functionality using a signal-slot paradigm. Each module holds a vector of function pointers. Each function pointer can be bind to a member functions of an object in the subsequent layer. Each module call the vector of functions in sequence to send through their processing results as the input of subsequent functions. 
This means each piece of sensor data will go through a depth first traverse of every module of a connected tree or pipeline.

The localiser provides not only the vehicle's global and local poses and their covariance matrix, but also update statistics such as update innovation, confidence, and status of outlier rejection. These real-time localiser performance metrics are monitored and used by the BT to make informed decisions. 
\begin{figure}
    \centering
    \includegraphics[width=\columnwidth]{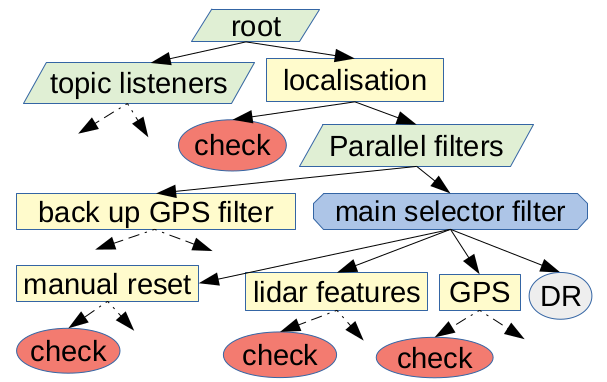}
    \caption{The behavior tree used in experimental evaluation on the University of Sydney campus dataset. Due to the complexity of the BT, lower level nodes that are explained in the methodology in section \ref{bTree} are omitted.}
    \label{fig:BT}
\end{figure}

In Fig. \ref{fig:localiser} we show one example of a pipeline construction. We use green squares to denote input/output interfaces of each module and black lines to connect them together. A Rosbag reader loads GPS, lidar, IMU, and encoder measurements from a Rosbag. These measurements are fed into a GPS observation model, lidar feature extraction module, and 2D motion model respectively. The results of the second layer modules are then fed into an Extended Kalman filter(EKF). The results of the EKF are fed to the destination interface layer to be published to ROS environment, written to a GIS database, and also stored in a rosbag.

% for example, the output of both gps and lidar sensor models are formated to be SE2(2D pose) type, so that they can be both accepted as input of SE2 Kalman Filter module in the subsequent layer. Equally, the output of GPS sensor model can be fed into both Kalman filter and a local optimiser, since they both take SE2 as input and operate in the SE2 space.

\section{The context aware behavior tree(BT) framework}\label{bTree}
We assume the readers are familiar with how BTs work. Please see section \ref{bt_intro} for a concise introduction to BTs. In Fig. \ref{fig:bt_models}, we propose examples of BT usages for autonomous vehicle localisation. In the following text sections, we use text enclosed in double quotation marks to denote a BT node. 

\subsection{Information gathering}
In order to monitor the performance of sensors and algorithms, the BT is first required to gather the live stream of sensor and localiser performance metrics. As seen in Fig. \ref{fig:bt_models}(a), the parallel ``root" node of BT always executes the ``topic listeners" node at the beginning of every tick before any decision making processes and action nodes. The ``topic listeners" subtree takes a snapshot of inputs at the beginning of a BT traverse, so that during the subsequent execution of BT, the same set of input data are used to make consistent decisions in all subsequent nodes. Having such a mechanism is necessary because multi sensor data arrives asynchronously and in high frequency. 
% A second thread(In our implementation this mechanism is handled by ROS subscribers) collects a fixed length cache of sensor streams as soon as they come in. When a listener node is ticked, the new information collected in the interval between current tick and the previous tick are moved to the blackboard of BT to be used by subsequent decision making tasks. the listener nodes execute in parallel under the "topic listeners" parallel node, because they collect independent streams of data such as measurements from different sensors.

\subsection{Coordinate state transitions for a single sensor modality} \label{singleSensor}
Fig. \ref{fig:bt_models}(b) provides an example subtree for one localisation estimation step using a single sensor and feature type. This subtree can be used to coordinate and supervise a single sensor modality, or can also serve as building blocks in multi sensor localisation. We use sensor A to represent a template sensor modality. 
During online localisation, the subtree monitors, commands and reconnects a data processing pipeline in the localiser depending on the situation and environmental changes. The 3 subtasks, namely ``initialise", ``localise" and ``fail" are executed in sequence. 

\subsubsection{initialisation}
Upon insertion of the ``sensor A localisation" subtree into the main BT, the BT gives commands to localiser to construct the necessary front end modules to attempt initialisation. In the example of lidar feature localisation, modules are connected together to extract features from point cloud, and match features to a prior feature map. The first child node of ``initialisation" checks if initialisation criteria are met in the front end of sensor A. As long as criteria are not met, the ``check" node will return a running state to denote sensor A is still in progress of initialisation. During this period the rest of sensor A subtree is blocked from being executed. Once the check succeeded, execution pass on to the following nodes to set up both BT utilities and connect the front end to the rest of modules in the localiser. 
% The sensor A measurements are processed by and flow through modules in all 4 layers in the localiser. 

\begin{figure}
    \centering
    \includegraphics[width=\columnwidth]{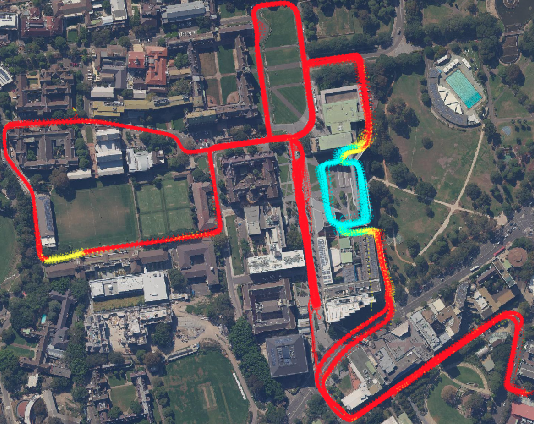}
    \caption{Vehicle trajectories of 53 rosbags in the University of Sydney Campus Dataset output from the backup GPS localisation subtree is superimposed together onto a satellite image. The color of trajectory points indicates the state of update used by the filter. Red: GPS updating. Yellow: GPS measurement rejected. Blue: GPS no fix. GPS measurements are consistently rejected at the entrance/exit of the underground carpark, and a narrow treetop covered section of road at the left of the image.}
    \label{fig:gps_traj}
\end{figure}

\subsubsection{localisation}
Having finished its initialisation sequence, sensor A localisation enters the ``localise" selector node and start the localisation process.

Current sensor fusion solutions often make assumptions about Gaussian distributed sensor noise profiles that do not change while driving. This is not true in urban environments and GPS violates this assumption in some areas worse than others. To overcome this issue, we encoded BT with a variable rejection method that is more restrictive when the position estimation is good, and accepts more outliers when the position estimation is already bad. In addition, this is a good example of where a separation of decision making from the data processing makes the code structure much more modular and manageable.

In the example of Fig. \ref{fig:bt_models}(b), to robustify the outlier rejection algorithm, we designed a 3 step fall back strategy. 
% so that a stricter or more relaxed policy for outlier rejection can be used according to better or worse localisation performance. 
Three children nodes range from more restricted, to more relaxed outlier rejection criteria: incorporating sensor measurements inside \(\pm2\) standard deviation(``2\(\sigma\)") of the localiser estimated distribution, \(\pm3\) standard deviation(``3\(\sigma\)")  or incorporating all(``all"). The BT execution will stay in the first node``2\(\sigma\)" when localisation performs well, and a 2\(\sigma\) bound is used in the sensor A pipeline of the localiser. When localisation performance decreases, the BT will fall back to more relaxed rejection bound nodes of ``3\(\sigma\)" and ``all", so that more observations become inliners. If these inliners are correct updates, the localiser pipeline will recover to better performance. The ``localising" nodes monitor the localiser pipeline performance. If the performance criteria is satisfied, the localiser pipeline returns running and keep using the same rejection bound. Otherwise the ``localising" node returns failure. 
The 3 step fall back strategy gives the localiser a better chance to recover from incorrect estimations. As soon as localisation estimation has improved, the BT execution performs a 1-3 step recovery process to go back to the stricter rejection nodes on the left. 

\subsubsection{failure}
If localisation performance remains poor and all 4 branches of the ``localise" subtree return failure, the execution proceed to the ``fail" sequence, in which the localiser pipeline is disconnected and relevant BT data are cleaned up. In future ticks, the sensor A subtree returns to check quality of sensor A front end to see if initialisation sequence is permitted to run.

\subsection{Multi sensor selector}
Fig. \ref{fig:bt_models}(c). shows a fall back strategy to select the best available sensor modality from a on-board multi sensor suite. This sensor switching ability is essential for a robust online localisation program, because the sensor/feature availability and noise profile change a great deal in diverse environments. 

Subtrees for different sensor modalities are inserted under the selector from left to right according to higher to lower expected localisation accuracy. Though lower priority modes of localisation are less accurate, their performance are usually more consistent and less susceptible to environmental changes, such as DR using encoders. 

At the startup ticks of the ``sensor selector" node, the initialisation criteria of every child is checked from left to right, until one check passes and the corresponding localiser pipeline is enabled. The children to the right of the running child are skipped during this time, since a more accurate sensor pipeline is already running. The children to the left(higher priority) of the currently running subtree keep attempting to initialise and checking initialisation criteria.

Once a higher priority subtree passes its initialisation check, the running subtree disables its localiser pipeline, and a new pipeline is enabled by the higher priority subtree. When a running subtree fails, execution falls back to check if any of the lower priority subtrees pass initialisation. In this way, the selector keep selecting more accurate modes of localisation as soon as they become available, whereas when higher priority localisation fail in adverse situations, the failures are detected immediately and localisation falls back to use less accurate and more robust sensor modalities.

\subsection{Parallel multi hypothesis localisation}
Because of the scalable and modular structure of BT, keeping multiple parallel running localiser pipelines becomes possible, see Fig. \ref{fig:bt_models}(d). In addition to the main selector subtree that may be used by safety critical down stream tasks such as path planning and control, parallel running hypothesis subtrees can fail, be inserted or pruned without having to affect the main subtree.

\begin{figure}
    \centering
    \includegraphics[width=\columnwidth]{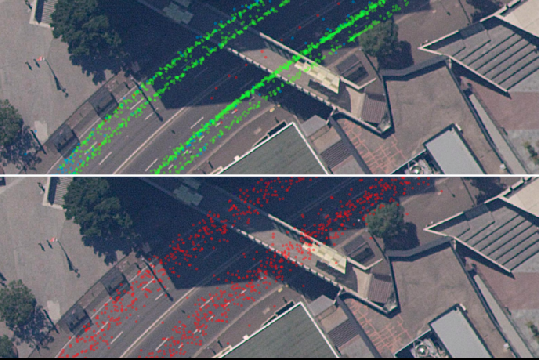}
    \caption{Zoomed in trajectories of 53 rosbags when vehicle drove on a 6 lane road. The main filter(top) achieved lane-level accuracy, whereas the GPS filtered trajectories(bottom) have a higher spread. The average translation standard deviation of GPS localising is 1.26m. Average lidar standard deviation is 0.25m.}
    \label{fig:lane_level_acc}
\end{figure}

% \subsection{discussion}\label{discussion}
% external sensors' noise profiles are a function of the environment. Many of the influence factors are spatial dependant. better outlier rejection using sensor measurement update history

% we only see the need of fusing multiple external sensors when ...
% Ideally we want to always incorporate all updates from all sensors, but this is only good when all sensors have zero mean, or sensor models and uncertainty distributions are characterised perfectly, and back-end is able to reject all outliers perfectly. In reality, we want to disable using less accurate modalities, because they most likely worsen the localisation result. To choose among a few mutually exclusive modalities, the most accurate modality is always the one we want. When the best accuracy modality is out, we fall back to less accurate ones. However, in a few situations the accuracy of sensor modalities changes, therefore our preference changes according to the relative accuracy changes. For example, when vehicle drive from outdoors to indoors, gps accuracy deteriorate and cannot be detected purely from GPS readings. GPS falsely report high accuracy readings and there is a latency before bad gps status is reported. 
% Another case is when we only lose best accuracy modality for a brief period/stretch of road, and updates resume after that. In this scenario, we gauge the necessity of changing priority 

\subsection{Using context information to facilitate decision making}\label{context}
The BT collects and monitors 2 sources of context information. The first is real-time localiser states and statistics to understand how the localiser is performing. This will detect localisation failure and performance changes so that sensor switching or lost recovery can be triggered. 

In addition to real-time localiser outputs, we compute location dependent sensor performance metrics as the second source of information. This information provide robustness against locations where tough environments greatly affect sensor performance, but the noisy sensor conditions are often not reflected in real-time sensor and localiser outputs.

An ideal localisation algorithm is expected to be capable of rejecting all bad measurements. However in reality, sensor noise profiles often change with respect to location. For example, when vehicle drive from an open road to a section of road in between tall buildings, the increased bias and possibility of multipath are usually not reflected by the GPS covariance matrix and the change of uncertainty remains undetected. 
It is very challenging to perform the correct updates and outlier rejections in both situations using the same generic sensor model. 
In our framework, the responsibility of detecting poor performance areas is partially shifted to a precomputed location dependant sensor model. We extract this information from a set of past localisation data when the location dependant noise was successfully detected, and store them in a map database. 
During online localisation, the BT monitors the location dependant sensor model and dynamically condition the localiser sensor model, so that they better suit the changing environment.

For example, if past localisation data shows sensor A is noisy at location B, the BT can detect this change of sensor noise profile by querying the sensor model database. The BT will make decisions to react to this change by rejecting more outliers or falling back to more robust sensor modalities. 

An example of GPS localisation is when vehicles are at entrance or exit of tunnels; it is usually very hard for a static localisation algorithm itself to correctly reject the outlier GPS observations. In a BT framework, BTs will check past GPS data to infer the quality of measurement at a certain location. Because the noisy measurements were detected at the entrance of a tunnel, BT will apply a stricter outlier rejection criteria at such locations and relax the criteria when vehicles move out of the noisy areas.

With the location dependant sensor models, it is also possible to predict the accuracy, availability and usability of different sensor modalities. The priority of sensors can be computed based on such information. When priorities have changed, the positions of subtrees in a BT can be dynamically rearranged to reflect the priority changes. In addition to priority changes, depending on the availability or usability of sensor modalities, sensor subtrees can also be inserted or pruned as vehicles move into different areas.

% In real time localisation, the higher level logical nodes may make a decision to insert, prune, or change priority of subtrees from the BT.  

\begin{figure}
    \centering
    \includegraphics[width=\columnwidth]{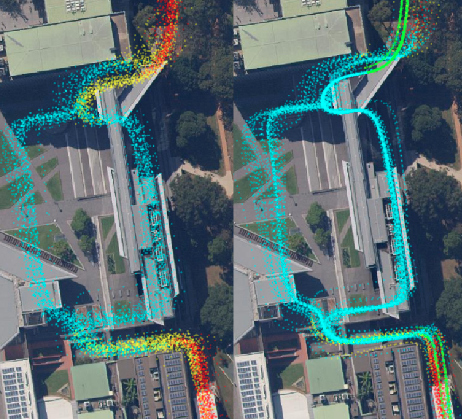}
    \caption{Zoomed in trajectories of 53 rosbags when vehicle drove through an underground carpark. Right: GPS+lidar main filter. Left: backup GPS filter}
    \label{fig:carpark}
\end{figure}

\section{EXPERIMENTAL EVALUATION} \label{experiments}

\subsection{Datasets}
A weekly drive was taken over a period of 1.5 years, following a consistent route that covered a wide variety of scene structures, including three-lane roads with high density traffic, large and small-scale loops, an underground car park, and shared pedestrian areas\cite{zhou2020developing}. Our dataset does not provide localisation ground truth, as RTK fix is not available in most areas of the map.
Among all the rosbags in the dataset, we evaluate the BT localisation framework on a set of 53 rosbags that do not contain sensor dropouts or systematic GPS biases more than 20m. Each rosbag recorded similar driving routes of around 20 minutes and 4km. 

\subsection{Building a geographically registered multi-sensor map database}
We built 3 layers of geographically registered sensor information and maintained them in a PostGIS database. The first layer contains the type and coordinates of pole-like and building corner features extracted from lidar pointclouds. This map layer is made and geo-registered following \cite{yi2019geographical} using motion corrected \cite{shan2020probabilistic} lidar pointclouds, GPS, IMUs and encoders measurements from the rosbag dated 2018-11-27. 
The second and third layers contain the location dependant sensor model of GPS and lidar respectively. The methodology of lidar feature localisation pipeline can be found in \cite{yi2019metrics}, and is therefore not repeated here.

% To generate the GPS spatial sensor model layer, a simple EKF using 95\% confidence bound in chi square test is used to filter through bags from September 2018 to January 2019. Among them, we selected 8 bags which the simple EKF is able to generate a mostly correct trajectory. For each GPS measurement in the selected bags, we store the vehicle pose estimation from EKF, whether GPS is able to resolve fix or no fix, and whether the measurement is an inliner or outlier.

% To generate the lidar features spatial sensor model layer, a lidar-GPS selector BT is used to filter through bags from June to August 2019. The GPS spatial sensor model layer is used to facilitate the GPS outlier rejection algorithm. Among the processed bags, we selected 6 bags which the localiser is able to generate a mostly correct trajectory. For each BT tick in the selected bags, we store the vehicle pose estimation, the ratio of converged versus not converged map matching frames of pointcloud features, and the ratio of inliner and outlier vehice pose observations outputted by the feature map matching algorithm, during the last tick interval. 

\subsection{BT localsation framework}
In our dataset, 10Hz encoder and 100Hz gyro are used for dead reckoning(DR), and 10Hz lidar pointcloud and 1Hz GPS are used for global updates using EKF filters. Fig.\ref{fig:BT} shows the BT used in the experiments. For simplicity, some structures that are already introduced in section \ref{bTree} are omitted. The BT contains two parallel running subtrees: a backup GPS filter and a main selector filter. We implemented the BT using the py\_tree\_ros package. The ``backup GPS filter", ``GPS" and ``lidar features" take the same form of section \ref{singleSensor} in which a selector node selects appropriate rejection bounds and decide whether or not to fall back to reset/reinitialise/fail actions.

The purpose of having a back up filter is that it serves as a fall back option when the main filter fails. The backup filter is also used as a reference to compare with the performance of the main filter. The ``reset" nodes implemented in the ``GPS" and ``lidar features" subtrees under main filter reset the main filter estimation to be the same as backup GPS estimation, when the heading difference between backup GPS filter and main filter are greater than 10 degrees and when backup GPS is better health state than main filter.
% The backup GPS filter does not have another localisation option to fall back to. Therefore the last fall back node of backup GPS triggers localiser pipeline to discard current estimations and reinitialise using upcoming GPS measurements. 

The main selector filter consists of 4 subtasks. The highest priority task is a manual reset node that monitors incoming reset messages from human input or a higher level control software. A manual reset will interrupt any of the other three modes of localisation and reset the main filter to a given pose or reinitialise it. The other three subtrees decide which sensors are used to update the main filter. The three modes are lidar+DR, GPS+DR, and DR only. GPS and lidar are not used concurrently because lidar mode is significantly more accurate than GPS in our sensor setup.
% The "DR" node is the default option upon startup or the last fall back node of the main filter. 

% In addition to be a fall back mode of "lidar feature" localising, the "GPS" subtree also serves as one of the initialisation actions. Once "GPS" has been initialised, lidar feature front end will be able to use the GPS filtered output to bootstrap lidar feature to map matching, since lidar map is geographically registered.

\begin{figure}
    \centering
    \includegraphics[width=\columnwidth]{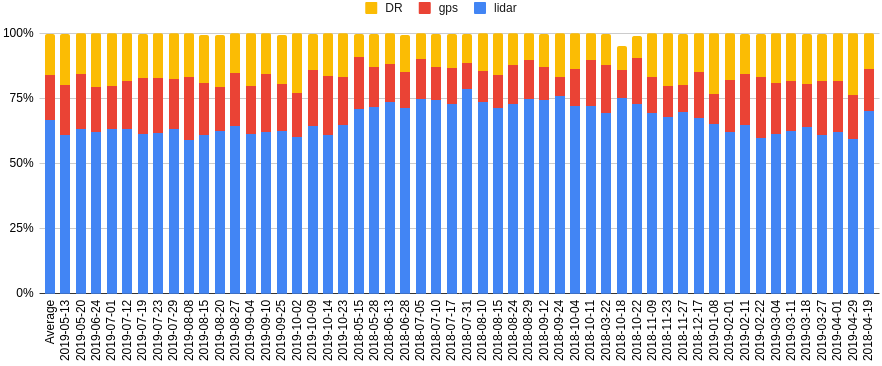}
    \caption{Percentage of distance traveled using blue: lidar feature to map matching updates; red: GPS updates; and yellow: no updates, only dead reckoning using gyro and wheel encoders. The average percentage of lidar, GPS and DR only in all 53 bags are 67\%, 17\%, and 16\%.}
    \label{fig:sensor_percentage}
\end{figure}

A visual demonstration of the working BT transitions and output of both filters are shown in the video attachment of this paper.

\subsection{Using location dependant sensor models to facilitate outlier rejection}

Once the BT starts receiving GPS measurements, the BT retrieves surrounding historical GPS measurements as well as current GPS health status to decide if GPS updates can be enabled. If so, both ``back up GPS filter" and the ``GPS" nodes are inserted to the BT. The resulting BT structure is shown in Fig. \ref{fig:BT}. On the other hand, at locations where the history GPS measurements show bad GPS performance, both ``back up GPS filter" and ``GPS" subtrees are pruned from their parent nodes. 

The location dependant lidar model is also used to facilitate lidar outlier rejection. To be an inliner of the filter, the output vehicle pose of map matching need to be within +- 10 degrees heading difference from a history vehicle pose retrieved from 10m box proximity from the location dependant lidar model database.

\section{RESULTS} \label{result}

\begin{figure}
    \centering
    \includegraphics[width=\columnwidth]{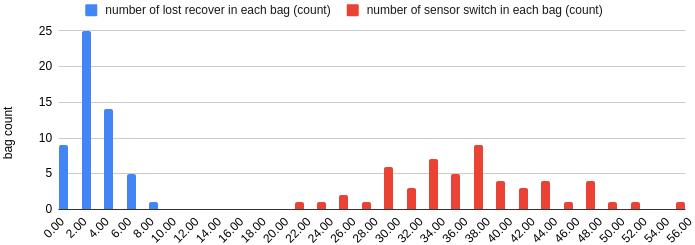}
    \caption{Number of lost recovery and sensor switching actions triggered in each of the 53 bags plotted as histograms. blue: lost recovery. red: sensor switching. On average, sensor switching happened 36 times and loss recovery was required 3 times per bag.}
    \label{fig:bag_count_jumps}
\end{figure}

The BT framework was run on all rosbags, representing 1.5 years of data. Fig.\ref{fig:gps_traj} shows the 53 trajectories of backup GPS filter outputs and the trajectories of main selector filter in Fig. \ref{fig:lidar_traj}. The trajectory points are drawn for every BT tick(500ms intervals). From these 2 pictures we can see the performance of both gps and lidar feature matching varies as a function of location, because the sensor transition patterns are consistent at a same location over the 1.5 years. The top image of Fig. \ref{fig:transition_zoomin} shows a zoomed view of a consistent sensor transition pattern at one location of the map. In all of the 53 bags, the BT chose to use lidar features at the feature rich region on the right(green) of Fig. \ref{fig:transition_zoomin}, transitioned to DR in the middle section of the road where features are scarce(blue), then switched to use GPS updates(red), and finally back to using lidar features again at the left of the figure. The bottom image shows the trajectories output from the backup GPS filter, where the vehicle positions are more spread out compared to the lidar+GPS main filter, indicating a higher localisation uncertainty compared to main filter. 

Fig. \ref{fig:lane_level_acc} shows the accuracy comparison between lidar based(top) and GPS based(bottom) filters. Lidar trajectories achieved lane level accuracy, whereas GPS trajectories did not.
Fig. \ref{fig:carpark} shows a zoomed in version of a challenging underground carpark area. Inside this underground carpark, both GPS and lidar localistion are unable to fix their poses. Using the context aware BT framework, the backup GPS filter is able to reject noisy measurements at entrance and exit of the carpark and use DR during underground driving (left image). Because of the presence of lidar features, the main filter is also able to switch to the more accurate lidar mode at the entrance and exit of the carpark (right image).

The percentage of distance travelled using lidar+DR, GPS+DR, and DR only in each bag are shown in Fig. \ref{fig:sensor_percentage}. The average percentage of lidar, GPS and DR only in all 53 bags are 67\%, 17\%, and 16\%. The average translation standard deviation of GPS localising is 1.26m. Average lidar standard deviation is 0.25m. This is expected according to the specification of the sensors. The average relative translation difference between backup GPS filter and main filter is 1.86m. During lidar localising, an average of 42.7\% of pointcloud feature frames are used for update, 12.6\% are rejected, and 44.7\% did not converge, among all the lidar scans in the 53 rosbags.

\begin{figure}
    \centering
    \includegraphics[width=\columnwidth]{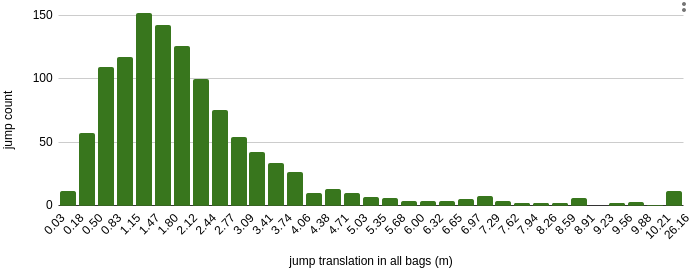}
    \caption{The jump or discontinuation distance of position estimation caused by both lost recovery and sensor switching actions. Jump distances from every triggered action in all of 53 bags are plotted as histogram. The average jump distance is 2.37m.}
    \label{fig:jump_translation}
\end{figure}

\begin{figure}
    \centering
    \includegraphics[width=\columnwidth]{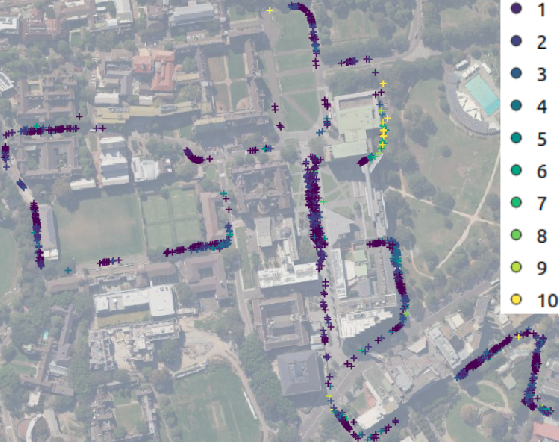}
    \caption{Locations on the map where each sensor switching and lost recovery action was triggered. The resulting jump distance of 0-10m is mapped to color using viridis colorramp. }
    \label{fig:transition_locations}
\end{figure}

On average, sensor switching happened 36 times and loss recovery was required 3 times per bag. Fig. \ref{fig:bag_count_jumps} shows a more detailed histogram of how many times the 2 types of events happened in the 53 bags. These events usually result in a discontinuity or jump in the filter pose estimation. The average translation magnitude of the relative pose or jump distance of all events in all bags is 2.37m. A distribution of average jump distances in each of the 53 bags are detailed in Fig.\ref{fig:jump_translation}. The jump difference of 1-3m is in accordance with the difference of localisation uncertainties between lidar and GPS modes. Fig.\ref{fig:transition_locations} shows the positions of individual events where color denotes the magnitude of jump distance.

\begin{figure}
    \centering
    \includegraphics[width=\columnwidth]{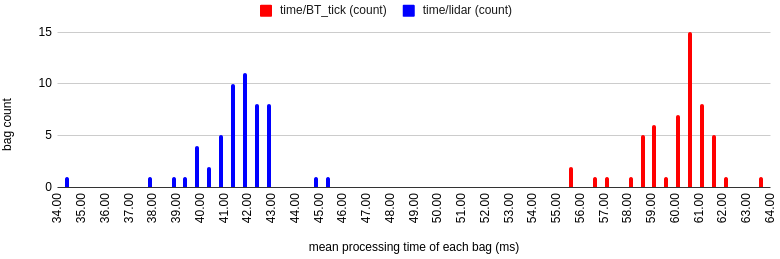}
    \caption{Red: Histogram of average processing time of each BT tick in 53 bags. Blue: Histogram of average processing time of each lidar pointcloud processed in 53 bags. The average time consumed per BT tick is 60.17ms(in python), per lidar pointcloud is 41.44ms(C++)}
    \label{fig:mean_time}
\end{figure}

\begin{figure}
    \centering
    \includegraphics[width=\columnwidth]{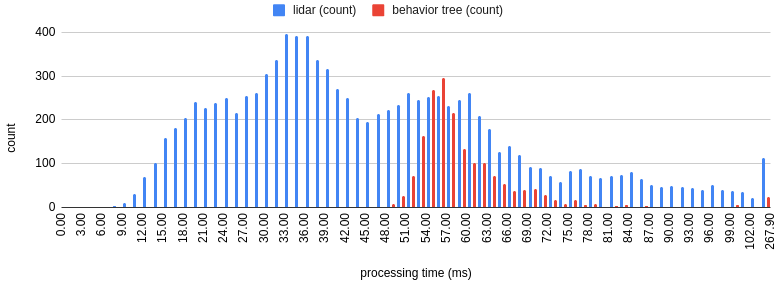}
    \caption{Red: Histogram of processing time of all BT ticks in one bag(2018-11-27). Blue: Histogram of processing time of all lidar pointclouds processed in the same bag.}
    \label{fig:time_week38}
\end{figure}

The average time consumed per BT tick is 60.17ms(in python), per lidar pointcloud (from feature extraction to pose distribution output) is 41.44ms (C++), and per GPS measurement (going through both the back up filter and the main filter) is 0.15ms. A detailed distribution of mean BT and lidar processing time for each bag is shown in Fig.\ref{fig:mean_time}. A histogram of processing time for all lidar pointclouds and all BT ticks in bag dated 2018-11-27 is shown in Fig.\ref{fig:time_week38}.

\section{CONCLUSIONS AND FUTURE DIRECTIONS}

In this paper, we proposed a context aware behavior tree(BT) localisation framework. This framework is able to monitor the health of real-time localisation output, and the noise profile of multiple localisation sensor modalities. The framework reacts to the situation and context changes by making decisions to perform intelligent behaviours such as sensor switching and loss recovery. This methodology was evaluated and achieved persistent localisation in a large and diverse urban environment over a time period of 18 months.

Due to the modular and flexible nature, the use of the behavior tree frame work is not confined to GPS and lidar feature combined EKF system as described in this paper. In the future, we would like to further robustify the framework by integrating additional sensor modalities such as RTK-GPS and visual features.

Instead of enacting localisation policies using handcrafted BT, we would also like to explore methods that automatically generate BTs. Reinforcement learning and evolutionary algorithms can be used in the future to generate BT policies to adapt to more diverse environments and train transitioning parameters using large amount of driving data.

% long term mapping: during the 20 months of data, the environment changed a lot, but the map is kept the same.

% sensor noise profile often change with respect to context such as vehicle motion and time. 

% Active localisation: localisation actions that can achieve best accuracy/robustness given a planned trajectory. localisation actions that can achieve given a planned trajectory and accuracy/robustness. 

\bibliographystyle{ieeetr}
\bibliography{citation}

\addtolength{\textheight}{-12cm}   % This command serves to balance the column lengths
                                  % on the last page of the document manually. It shortens
                                  % the textheight of the last page by a suitable amount.
                                  % This command does not take effect until the next page
                                  % so it should come on the page before the last. Make
                                  % sure that you do not shorten the textheight too much.

\end{document}